\documentclass{article}

\usepackage{placeins}
\usepackage{adjustbox}
\usepackage{wrapfig,lipsum,booktabs}

\usepackage[preprint]{neurips_2025}

\usepackage[utf8]{inputenc} 
\usepackage[T1]{fontenc}    
\usepackage{hyperref}       
\usepackage{url}            
\usepackage{booktabs}       
\usepackage{amsfonts}       
\usepackage{nicefrac}       
\usepackage{microtype}      
\usepackage{xcolor}         
\usepackage{graphicx}       
\usepackage{subcaption}     
\usepackage{float}          

\usepackage{changepage} 

\title{A Women's Health Benchmark for Large Language Models}

\newcommand{\hs}{-0em}

\author{
\hspace{\hs}\textbf{Victoria-Elisabeth Gruber}$^{1}$ \quad
\textbf{Razvan Marinescu}$^{1}$  \quad
\textbf{Diego Fajardo}$^{1}$ \quad
\hspace{\hs}\textbf{Amin H. Nassar}$^{2}$ \\
\textbf{Christopher Arkfeld}$^{3}$ \quad
\textbf{Alexandria Ludlow}$^{4}$ \quad
\hspace{\hs}\textbf{Shama Patel}$^{5}$ \quad
\textbf{Mehrnoosh Samaei}$^{6}$ \\
\textbf{Valerie Klug}$^{7}$\quad \hspace{\hs}\textbf{Anna Huber}$^{7}$ \quad
\textbf{Marcel Gühner}$^{7}$ \quad
\textbf{Albert Botta i Orfila}$^{7}$ \\
\hspace{\hs}\textbf{Irene Lagoja}$^{7}$ \quad
\textbf{Kimya Tarr}$^{8}$ \quad
\textbf{Haleigh Larson}$^{9}$ \quad
\textbf{Mary Beth Howard}$^{10}$ \\
\hspace{\hs}$^1$Lumos AI \quad
$^2$Medical Oncology, Yale Cancer Center\\ 
\hspace{\hs} $^3$Obstetrics and Gynecology, MGH, Harvard Medical School \\
\hspace{\hs}$^4$Obstetrics, Gynecology \& Reproductive Sciences, UCSF\\ 
\hspace{\hs}
$^5$Brown Division of Global Emergency Medicine \\
\hspace{\hs}$^6$Department of Emergency Medicine, Emory University\\
\hspace{\hs}$^7$Pharmacy Department, Clinic Ottakring \\
\hspace{\hs}$^8$Windrush Surgery, Buckinghamshire, Oxfordshire and \\ Berkshire West Integrated Care Board, NHS \\ \quad
\hspace{\hs}$^9$Women’s Health Research, Yale School of Medicine \\
\hspace{\hs}$^{10}$Johns Hopkins University School of Medicine \\ \quad
\hspace{\hs}\texttt{\{victoria,razvan,diego\}@thelumos.ai} \\
}

\begin{document}
\maketitle

\begin{abstract}

  As large language models (LLMs) become primary sources of health information for millions, their accuracy in women's health remains critically unexamined. We introduce the Women's Health Benchmark (WHB), the first benchmark evaluating LLM performance specifically in women's health. Our benchmark comprises 96 rigorously validated model stumps covering five medical specialties (obstetrics and gynecology, emergency medicine, primary care, oncology, and neurology), three query types (patient query, clinician query, and evidence/policy query), and eight error types (dosage/medication errors, missing critical information, outdated guidelines/treatment recommendations, incorrect treatment advice, incorrect factual information, missing/incorrect differential diagnosis, missed urgency, and inappropriate recommendations). We evaluated 13 state-of-the-art LLMs and revealed alarming gaps: current models show approximately 60\% failure rates on the women's health benchmark, with performance varying dramatically across specialties and error types. Notably, models universally struggle with "missed urgency" indicators, while newer models like GPT-5 show significant improvements in avoiding inappropriate recommendations. Our findings underscore that AI chatbots are not yet fully able of providing reliable advice in women's health.

\end{abstract}

\section{Introduction}

Artificial intelligence (AI) has made significant progress in various domains, including healthcare, with the potential to significantly improve patient care and quality of life \cite{patil2023AI} \cite{OLAWADE2024AI}. Medical diagnosis is one application area where AI has already shown promising results, as algorithms are able to identify patterns in  medical images and text that are not readily visible to the human eye \cite{pinto2023artificial}. Another field where AI is making significant progress is personalized medicine, where algorithms are identifying patterns in vast volumes of patient data and providing personalized treatment plans taking into account the patient’s unique genetics, lifestyle and medical history \cite{johnson2021precision}. AI is also being utilized to increase the effectiveness of clinical operations, through integration of chatbots and virtual assistants, to help with patient intake and triage \cite{alowais2023revolutionizing}. Hence, it is not surprising that more and more patients and physicians are turning towards AI chatbots for health related questions and get help in clinical decision making everyday. Roughly one in six adults has used an AI chatbot for health related questions in the last year \cite{kff_ai_health_info_2024}. This gives rise to the question: How accurate are AI models when it comes to women’s health-related questions?

Historically, women have been underrepresented in research and clinical trials, meaning that a majority of available data is biased toward male populations due to a myriad of reasons \cite{karpel2025missing} \cite{rittenberg2025women}. Some of the major reasons include biological sex differences (physiological differences driven by chromosomal, hormonal, and anatomical factors), gender effects (differences arising from socially constructed roles, behaviors, and identities) and inadequate research in women, such as excluding pregnant women and women of child-bearing age from participating in clinical studies for decades\cite{bartz2020clinical} \cite{national2024advancing} \cite{iom_sex_matter_2001}.
Bias in the training data of large language models arises not only from the historical underrepresentation of women in research, but also from shortcut learning and spurious correlations present in imbalanced or noisy medical training datasets. As a result, models may rely on simplified patterns or stereotypes instead of true medical reasoning, leading to confident but incorrect responses that can worsen sex- and gender-related gaps in healthcare.\cite{dogra2024shortcut} \cite{ye2024cleverhans} \cite{geirhos2020shortcut}
A recent study examining online reproductive health misinformation across multiple social media platforms and websites found that 23\% of the content included medical recommendations that do not align with professional guidelines \cite{john2025online}. They further found that potentially misleading claims and narratives about reproductive topics like contraception, abortion, fertility, chronic disease, breast cancer, maternal health, and vaccines are abundant across several social media platforms and websites \cite{john2025online}.  While some benchmarks exist for general healthcare, to our knowledge, no specific benchmark currently exists that evaluates the credibility and safety of AI models in the domain of women’s health.

In this paper, we investigate the accuracy of LLMs for women’s health-related questions. Our proposed benchmark consists of 5 specialties: obstetrics and gynecology, emergency medicine, primary care, oncology, and neurology. We run three types of queries: patient queries, clinician queries, and evidence/policy queries, as well as eight error types: (1) dosage/medication errors, (2) missing critical information, (3) outdated guidelines/treatment recommendations, (4) incorrect treatment advice, (5) incorrect factual information, (6) missing/incorrect differential diagnosis, (7) missed urgency, and (8) inappropriate recommendations. This is to ensure that the benchmark is comprehensive and covers a wide range of women’s health topics. We benchmarked the following models: Claude 4.0 Opus, Claude 4.0 Sonnet, Gemini 2.5 Flash, Gemini 2.5 Pro, Gemini 3 Pro, GPT-4o Mini, GPT-5, GPT-5.1, o3, o3 Mini, Grok 4, Ministral-8B, and Mistral Large. We found that the overall best performing model was GPT-5, followed by Gemini 3 Pro with best performance in obstetrics and gynecology. Importantly, no model demonstrated consistently high performance throughout all specialties, error types and query types.

An overview of our contributions is as follows: 

\begin{itemize}
  \item The Women’s  Health Benchmark (WHB) which includes 96 realistic open-ended prompts across five medical specialties (obstetrics and gynecology, emergency medicine, primary care, oncology, and neurology).
  \item It was produced with the help of 17 women’s health experts including clinicians, pharmacists, and researchers across the United States of America and Europe.
  \item We measure the performance of the WHB across 13 LLMs.
  \item Women’s health experts were asked to prompt from either a patient or doctor perspective or ask evidence-based questions.
  \item We release the WHB data and code via The Lumos AI Labs Hugging Face repository.
\end{itemize}

\section{Related Work}

A prime example of a field in which sex and gender differences have been extensively studied in recent years is cardiovascular disease. Cardiovascular disease has been perceived as a 'man's disease', and this misconception has led to under-diagnosis and treatment for women worldwide \cite{schenck2009risk}. In a recent study Burgess et al. evaluated the representation of women in cardiovascular randomized controlled trials and reported that women are significantly under-represented in clinical trials both as participants and researchers \cite{burgess2025under}. Another study observed that women are less likely than men to receive diagnostic tests, interventions, or appropriate preventive treatments, pointing to bias in diagnostic and care protocols \cite{al2024gender}.

Medical large language model benchmarks encompass a broad spectrum of tasks, including clinical knowledge representation, differential diagnosis generation, complex medical text summarization, and the emulation of clinical reasoning and empathy in patient interactions \cite{singhal2023expertlevelmedicalquestionanswering} \cite{mcduff2023accuratedifferentialdiagnosislarge} \cite{van_veen_2024} \cite{savage2023diagnosticreasoningpromptsreveal}. Another recognized benchmark in the field, HealthBench, advances the field by assessing clinician–patient interactions through question-specific evaluation rubrics \cite{arora2025healthbenchevaluatinglargelanguage}. One of the key challenges in AI in healthcare is the need to ensure that the AI systems are able to learn from data in a way that is not biased and that is able to generalize to new patients and new data. 

\begin{figure}[H]
\centering
\includegraphics[width=0.76\textwidth]{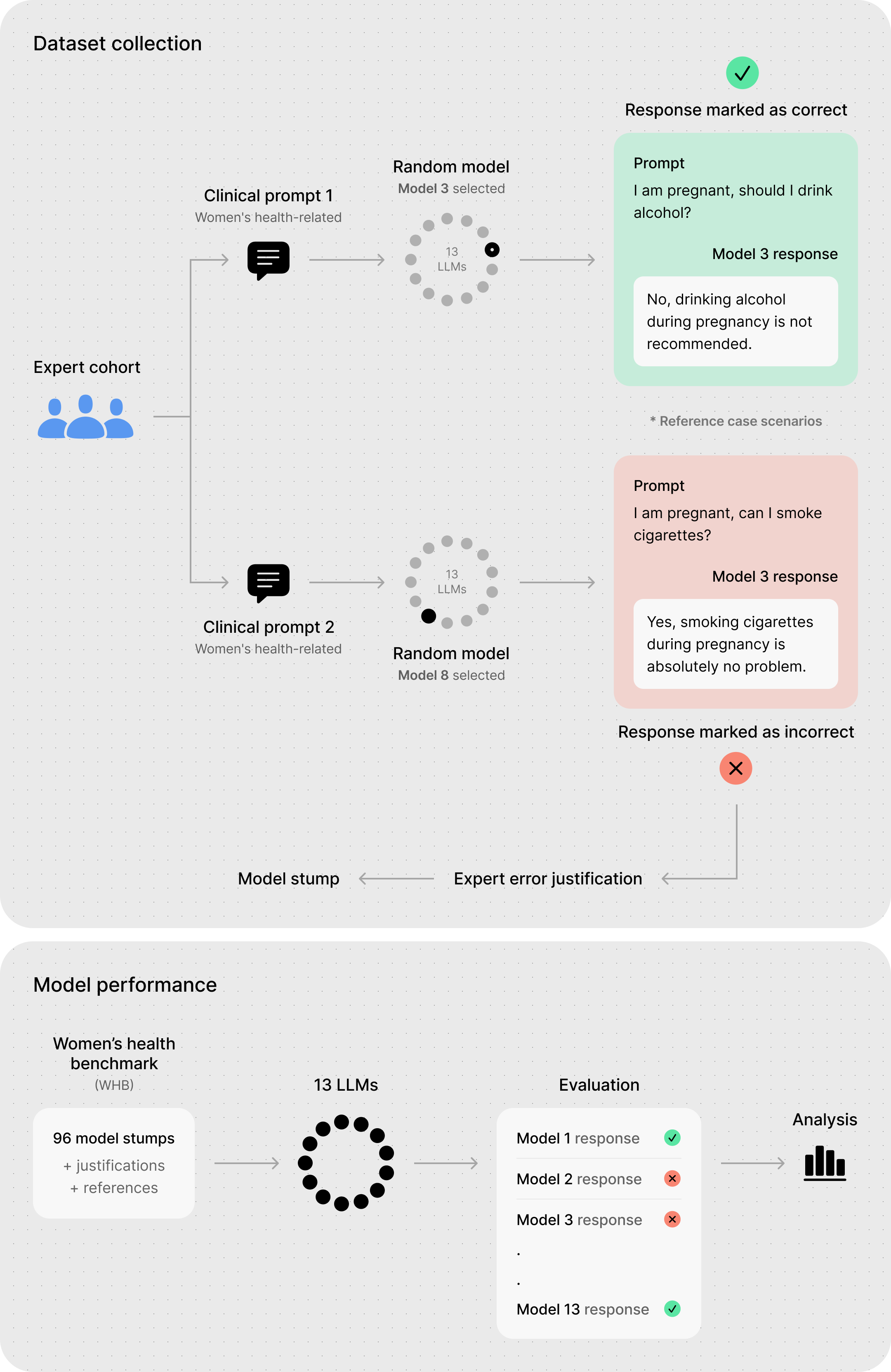}
\caption{Overview of WHB methodology. Top: Dataset collection workflow showing how the expert cohort generated women's health-related clinical prompts, which were randomly assigned to one of the 13 LLMs to identify incorrect responses that became model stumps. Bottom: Model performance evaluation workflow showing how the 96 model stumps were used to benchmark all 13 LLMs.}
\label{fig:whb_overview}
\end{figure}

\section{Methods}

In Figure~\ref{fig:whb_overview} we show an overview of our women's health benchmark study. A group of experts prompted women's health related questions which were randomly assigned to one of the 13 LLMs included in this study. The experts evaluated the model response and either accepted or rejected it. If rejected, they needed to provide a justification plus reference to the source or truth. The rejected model prompt was then, after going through an approval process, added to the collection of model stumps. In total, 96 model stumps were generated and collectively referred to as the WHB dataset. The WHB was then used to benchmark 13 LLMs. 

\subsection{Data collection}

\paragraph{Women's health experts.}
The model stumps (prompts) in WHB were created by a group of 17 experts over a period of six weeks. The group of experts included physicians (6), pharmacists (5), medical researchers (5) and one nurse practitioner with practice and research experience across the United States of America, the United Kingdom, and the European Union. In the group, 67\% were attending physicians or independent practitioners, and 33\% were fellows. We vetted physicians for their participation in this benchmark dataset creation through a multi-step process. We screened publications on Pubmed and performed a thorough web-search to find women’s health experts. Physicians had to have a minimum of two peer-reviewed publications in a women’s health related field and/or practice experience of a minimum of 3 years with focus women's health. Pharmacists had to have a minimum of 1 year of practical clinical experience after finishing their studies. Researchers were enrolled in either a PhD program or postdoctoral fellowship in a women's health related program. The nurse practitioner had an experience of 9 years of working at the Department of Obstetrics, Gynecology \& Reproductive Sciences. The selected experts had to go through an educational webinar and the quality of their submissions was routinely reviewed by qualified personnel at Lumos AI.

\paragraph{Expert instructions.}

\begin{wraptable}{r}{5.5cm}
\centering
\caption{Overview of evaluated language models.}
\label{tab:models}
\begin{tabular}{ll}
\toprule
\textbf{Model} \\
\midrule
Claude 4.0 Opus (2025-05-14)  \\
Claude 4.0 Sonnet (2025-05-14)  \\
Gemini 2.5 Flash (Preview 05-20)  \\
Gemini 2.5 Pro (Preview 05-06)  \\
Gemini 3 Pro  \\
GPT-4o Mini (2024-07-18) \\
GPT-5 (2025-08-07) \\
GPT-5.1  \\
Grok 4 (0709) \\
Ministral-8B (Latest) \\
Mistral Large (Latest) \\
OpenAI o3 (2025-04-16) \\
OpenAI o3 Mini (2025-01-31) \\
\bottomrule
\end{tabular}
\end{wraptable}

All participating experts received a standardized instruction document outlining the process for developing and evaluating benchmark questions. Experts were instructed to generate realistic, women-specific clinical prompts reflecting real-world contexts in which care, presentation, or treatment may differ for women. They were encouraged to frame their questions from diverse perspectives, such as a patient seeking advice, a clinician considering management options, or an evidence-based reviewer referencing current literature. Experts did not know which model would be used to answer their prompts. Following model response generation, experts were asked to assess each output for factual accuracy, safety, and completeness, explicitly identifying errors, outdated or unsafe recommendations, and any critical omissions that could potentially lead to patient harm. For each identified issue, experts were required to provide a concise written justification (1–3 sentences) describing the nature of the error and, whenever possible, include a supporting citation or link to the authoritative source (e.g., peer-reviewed article, clinical guideline, or regulatory document) confirming the correct information.

\subsection{Models}
Table~\ref{tab:models} provides a comprehensive overview of the baseline models used in this paper. To generate our raw dataset of 96 model stumps, models were applied in a rotating fashion. In the next step, for benchmarking, each model was evaluated on the 96 WHB prompts.

\subsection{Dataset}

\FloatBarrier

\paragraph{Dataset preparation.}
 In total, experts prompted 345 questions, a randomly selected model generated an answer. Experts then evaluated if the model provided a correct answer or not. 249 (72.2\%) questions lead to an answer classified as correct by the experts, and 96 (27.8\%) were classified as incorrect, making up the raw dataset. Questions were manually generated and submitted by our expert group trained according Lumos AI’s standardized instructions. In total, experts contributed 143 model stumps accompanied by written justifications and verified sources of truth, of which 96 were approved following internal quality review. To ensure data integrity and conceptual alignment with the benchmark’s objectives, rigorous preprocessing was applied. This process included the removal of duplicate entries, prompts that did not meet the women-specific criteria, and instances where the expert justification did not reflect a factually incorrect model response or where the identified error lacked sufficient clinical relevance or potential for harm. 

\paragraph{Dataset organization.}

The dataset is organized into five categories (obstetrics and gynecology, emergency medicine, primary care, oncology, and neurology), which reflect areas of real-world health interactions each corresponding to a medical specialty. Each category contains a list of single-turn model stumps, each with a corresponding justification and reference. 

\paragraph{Types of model stumps.}
In the dataset, the model stumps are further organized into three query types: patient query, clinician query, and evidence/policy query.
Patient queries are designed to be from a patient's perspective, asking questions about their health and treatment options. Doctor queries are intended to be from a doctor's perspective, such as prompting a realistic patient vignette. Evidence/policy queries are formulated to ask questions about the latest research and best practices in the field. The dataset contains 51 patient queries (53.1\%), 33 clinician queries (34.4\%), and 12 evidence/policy queries (12.5\%) (see Figure~\ref{fig:query_type}).

\paragraph{Model stumps by medical specialties.}
The dataset consists of 41 obstetrics and gynecology questions (42.7\%), 24 emergency medicine questions (25\%), 17 primary care questions (17.7\%), 11 oncology questions (11.5\%), and three neurology questions (3.1\%) (see Figure~\ref{fig:specialty}).

\paragraph{Model stumps by error type.}
The model stumps in our dataset can be categorized into distinct error types based on the nature of observed model stumps. The most common category is “dosage/medication errors” (17 model stumps, 17.7\%), defined as instances where the model provided an incorrect drug dosage, frequency, duration, or selected an inappropriate medication. This is followed by "missing critical information" (16 model stumps, 16.7\%), which included responses that omitted essential clinical details required to answer the question safely or accurately. Next is “outdated guidelines/treatment recommendations” (13 model stumps, 13.5\%), consisting of answers that reflected superseded clinical practices or failed to incorporate current standard-of-care guidance. “Incorrect treatment advice” (13 model stumps, 13.5\%) was defined as recommendations that were inconsistent with evidence-based management despite not necessarily being outdated.
Further, 12 model stumps are categorized as “incorrect factual information” (12.5\%), which included responses that contained false or misleading information. The remaining categories are “missing/incorrect differential diagnosis” (12 model stumps, 12.5\%), defined as either failing to list key differentials or proposing diagnostically implausible ones, “missed urgency” (9 model stumps, 9.4\%), which included failures to recognize the need for urgent escalation, and “inappropriate recommendations” (4 model stumps, 4.7\%), which included recommendations that were not appropriate for the patient's situation (see Figure~\ref{fig:error_type}).

\newcommand{\scale}{1.0}

\begin{figure}[H]
    \centering\begin{subfigure}[t]{0.35\textwidth}
        \centering
        \includegraphics[width=\scale\textwidth, trim=0 60 0 0]{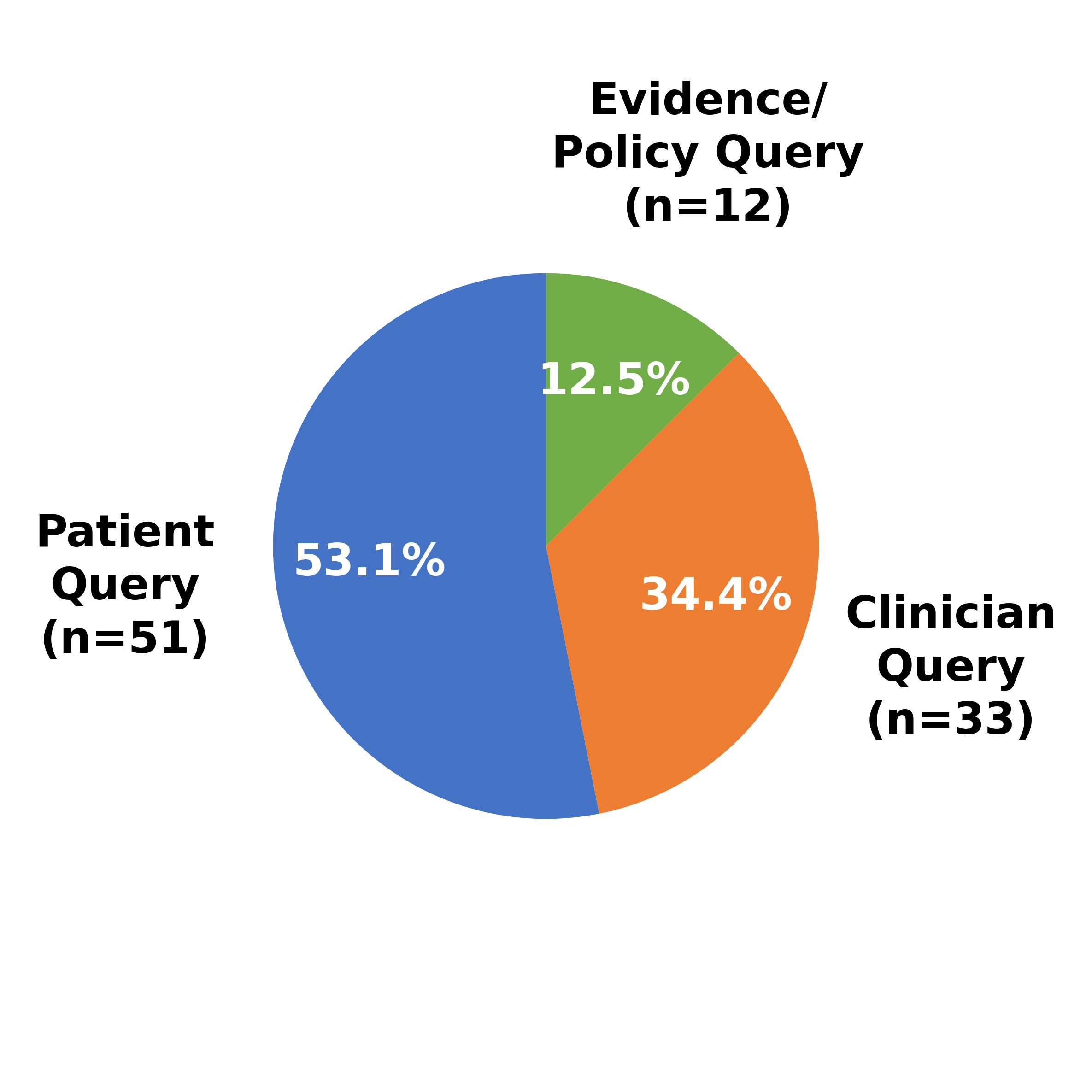}
        \caption{Query types}
        \label{fig:query_type}
    \end{subfigure}
    \begin{subfigure}[t]{0.4\textwidth}
        \centering
        \includegraphics[width=\scale\textwidth, trim=0 80 0 0]{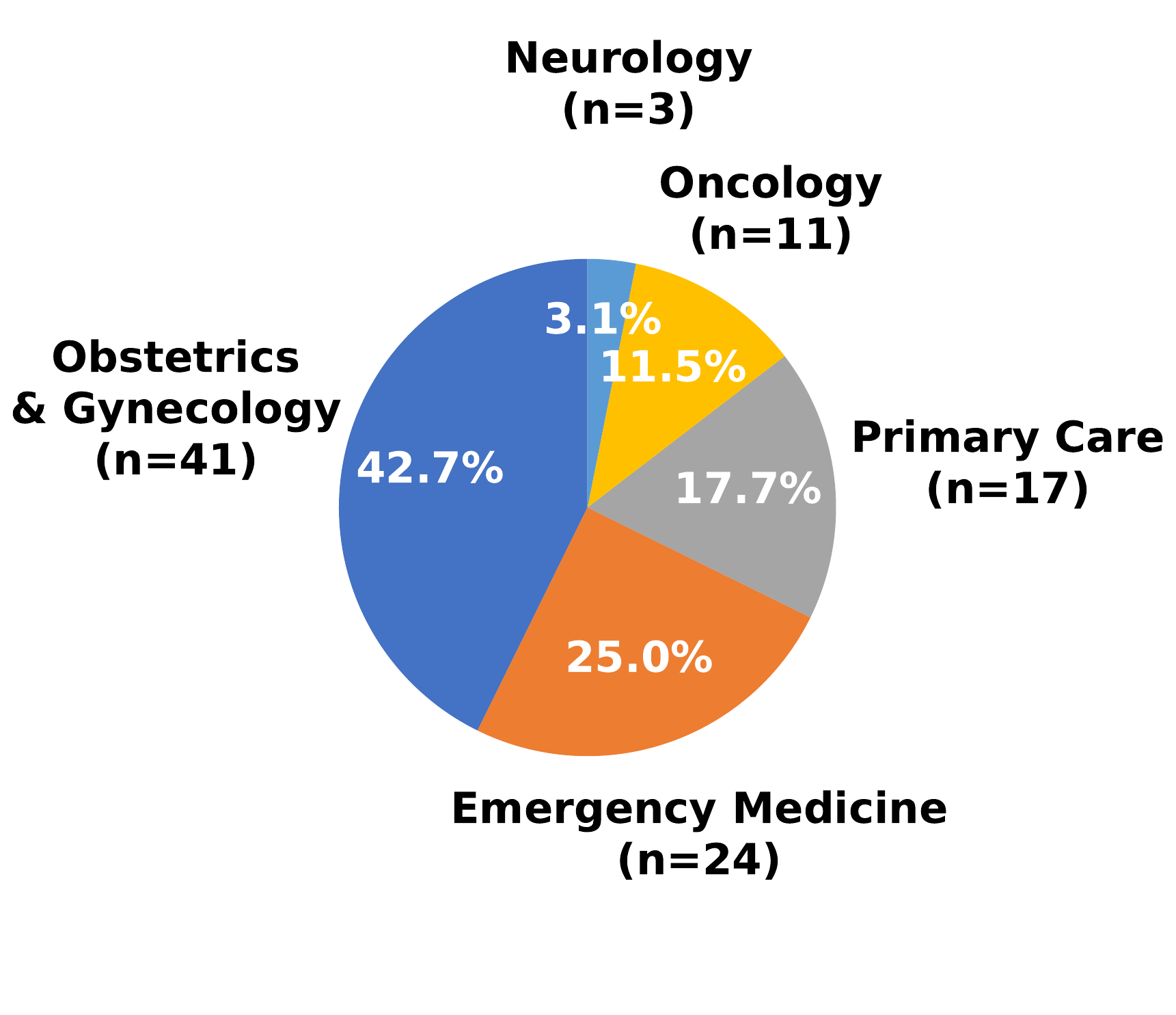}
        \caption{Medical specialties}
        \label{fig:specialty}
    \end{subfigure} 

    \vspace{2em}
    \begin{subfigure}[b]{0.6\textwidth}
        \centering
        \includegraphics[width=\scale\textwidth]{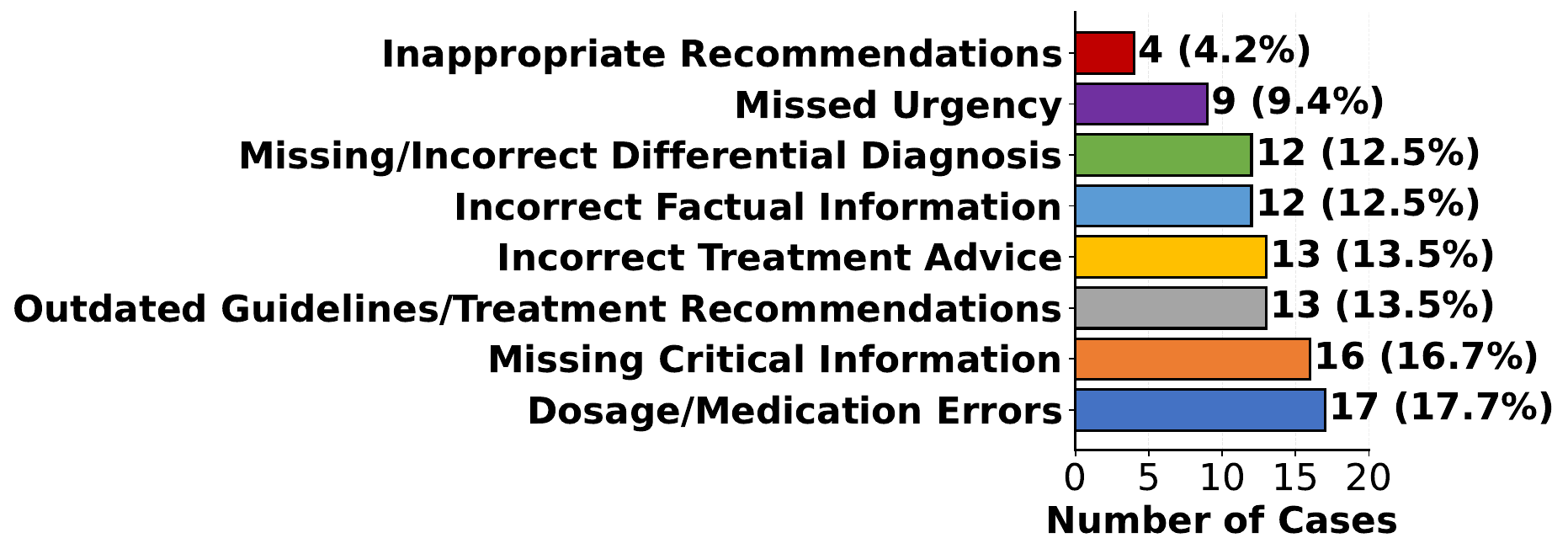}
        \caption{Error types}
        \label{fig:error_type}
    \end{subfigure}
    \caption{Distribution of model stumps in WHB. (A) Distribution by query type. (B) Distribution by medical specialty. (C) Distribution by error type.}
    \label{fig:distributions}
\end{figure}

\subsection{Human evaluation}
To evaluate the performance of the models on the WHB dataset, a human evaluator with a PhD in clinical sciences was used to assess the model answers to each stump. The evaluator screened the answers for the respective error that was identified by the expert group and either approved or rejected the answer if it included the identified error. The evaluator was only allowed to reject the answer if it included the exact same error as the one identified by the expert group, this does not mean that the approved answers are free of errors, but that the errors are not the same as the ones identified by the expert cohort.

\subsection{Definitions}

\begin{itemize}
\item \textbf{Model stump:} A model stump is a prompt that leads to an incorrect model answer.
\item \textbf{Case:} A case is a model stump plus a model answer (see figure 6 in appendix).
\item \textbf{Incorrect case:} An incorrect case is a case that was rejected by the human evaluator on grounds of the identified error by the expert group.
\item \textbf{Correct case:} A correct case is a case that was approved by the human evaluator on grounds of not identifying the exact same error as the one identified by the expert cohort.
\end{itemize}

\section{Results}

\subsection{Model performance}

\begin{figure}[H]
    \centering
    \includegraphics[width=\textwidth]{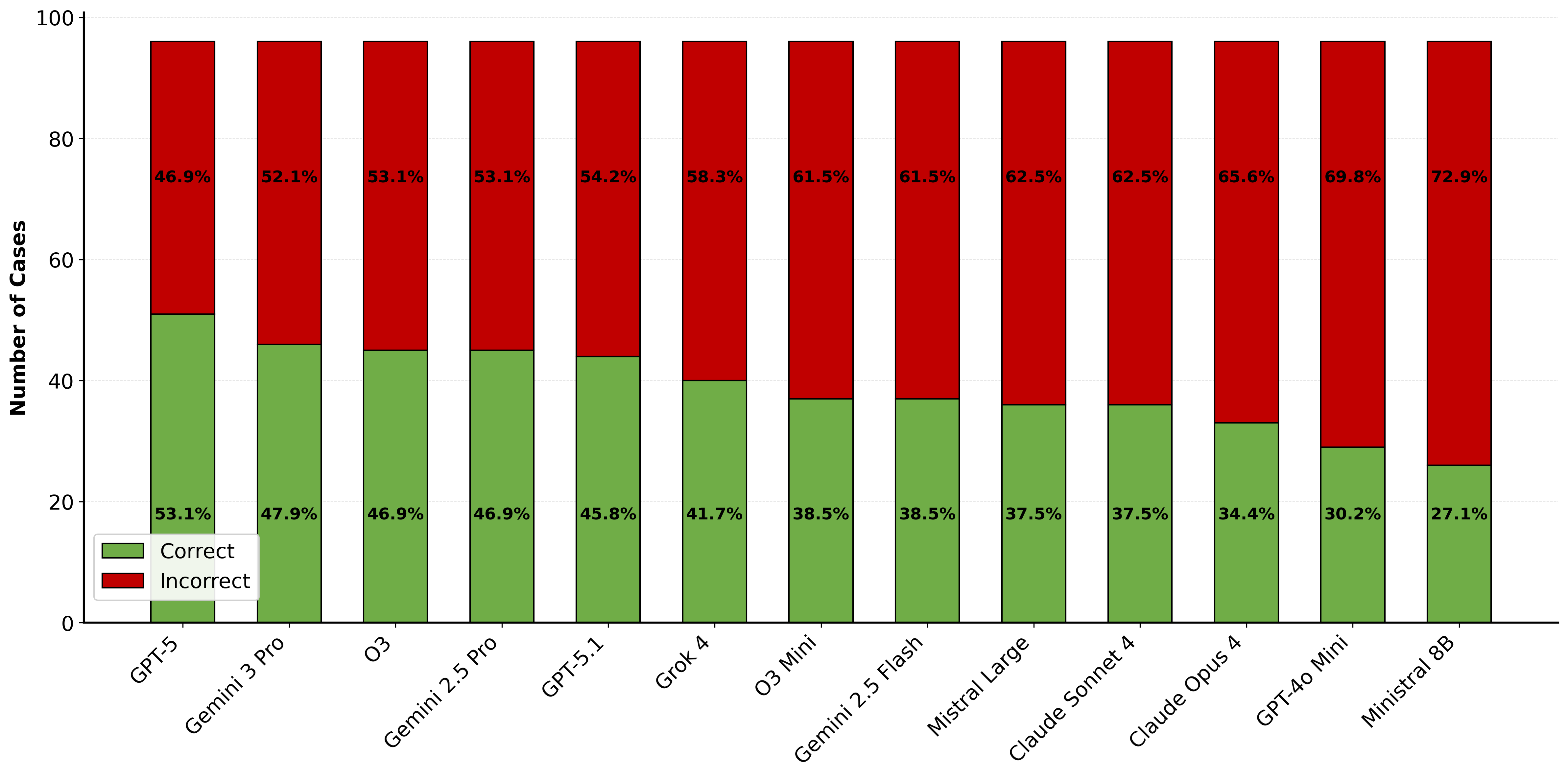}
    \caption{Model performance on WHB. The approval rate was calculated as the percentage of correct cases (correct / total number of cases), shown in green columns. The failure rate was calculated as the percentage of incorrect cases (incorrect/total number of cases), shown in red columns. }
    \label{fig:model_performance}
\end{figure}

Figure~\ref{fig:model_performance} shows for all analyzed LLMs the total number of correct and incorrect cases. As expected, larger models generally perform better than the smaller models. The mean approval rate for large LLMs (Claude Opus 4, Claude Sonnet 4, Gemini 2.5 Pro, Gemini 3 Pro, GPT-5, GPT-5.1, o3, Grok 4, Mistral Large) is 44\% and for small models (Gemini 2.5 Flash, GPT-4o Mini, Ministral 8B and o3 Mini) 34\%. The best performing model is GPT-5 with an approval rate of 53.1\%, followed by Gemini 3 Pro with an approval rate of 47.9\%, and o3 with an approval rate of 46.9\%. The worst performing model is Mistral 8B with an approval rate of 27.1\%. 

\paragraph{Model performance by query type.}
We analyzed the performance of the models on the three different query types: patient query, clinician query, and evidence/policy query. Table~\ref{tab:query_type_performance} shows the failure rates by model and query type. The failure rates were calculated as the percentage of incorrect cases (incorrect/total). Overall, GPT-5 shows the best performance on all query types, with a failure rate of 47.1\%, 51.5\%, and 33.3\% for patient query, clinician query, and evidence/policy query respectively. Ministral-8B shows the worst performance on all query types, with a failure rate of 78.4\%, 57.6\%, and 91.7\% for patient query, clinician query, and evidence/policy query respectively. Overall, the models perform nearly equally on patient and clinician queries. However, models like Grok 4, Ministral 8B, and Gemini 3 Pro show dramatically better performance on clinician queries compared to patient queries.

\begin{table}[H]
\centering
\caption{Failure rates by model and query type. Values show percentage of incorrect responses (incorrect/total) with 95\% confidence intervals. Bold shows the best values obtained in each metric.}
\label{tab:query_type_performance}
\adjustbox{max width=\textwidth}{%
\begin{tabular}{lccc}
\toprule
\textbf{Model} & \textbf{Patient Query} & \textbf{Clinician Query} & \textbf{Evidence/Policy Query} \\
\midrule
GPT-5 & \textbf{47.1\% (24/51) [34.1\%, 60.5\%]} & 51.5\% (17/33) [35.2\%, 67.5\%] & \textbf{33.3\% (4/12) [13.8\%, 60.9\%]} \\
Gemini 3 Pro Preview & 60.8\% (31/51) [47.1\%, 73.0\%] & \textbf{42.4\% (14/33) [27.2\%, 59.2\%]} & 41.7\% (5/12) [19.3\%, 68.0\%] \\
Gemini 2.5 Pro & 54.9\% (28/51) [41.4\%, 67.7\%] & 45.5\% (15/33) [29.8\%, 62.0\%] & 66.7\% (8/12) [39.1\%, 86.2\%] \\
OpenAI o3 & 49.0\% (25/51) [35.9\%, 62.3\%] & 60.6\% (20/33) [43.7\%, 75.3\%] & 50.0\% (6/12) [25.4\%, 74.6\%] \\
GPT-5.1 & 54.9\% (28/51) [41.4\%, 67.7\%] & 51.5\% (17/33) [35.2\%, 67.5\%] & 58.3\% (7/12) [32.0\%, 80.7\%] \\
Grok 4 & 68.6\% (35/51) [55.0\%, 79.7\%] & \textbf{42.4\% (14/33) [27.2\%, 59.2\%]} & 58.3\% (7/12) [32.0\%, 80.7\%] \\
Gemini 2.5 Flash & 56.9\% (29/51) [43.3\%, 69.5\%] & 63.6\% (21/33) [46.6\%, 77.8\%] & 75.0\% (9/12) [46.8\%, 91.1\%] \\
OpenAI o3 Mini & 62.7\% (32/51) [49.0\%, 74.7\%] & 66.7\% (22/33) [49.6\%, 80.2\%] & 41.7\% (5/12) [19.3\%, 68.0\%] \\
Claude 4.0 Sonnet & 66.7\% (34/51) [53.0\%, 78.0\%] & 60.6\% (20/33) [43.7\%, 75.3\%] & 50.0\% (6/12) [25.4\%, 74.6\%] \\
Mistral Large & 62.7\% (32/51) [49.0\%, 74.7\%] & 57.6\% (19/33) [40.8\%, 72.8\%] & 75.0\% (9/12) [46.8\%, 91.1\%] \\
Claude 4.0 Opus & 64.7\% (33/51) [51.0\%, 76.4\%] & 69.7\% (23/33) [52.7\%, 82.6\%] & 58.3\% (7/12) [32.0\%, 80.7\%] \\
GPT-4o Mini & 66.7\% (34/51) [53.0\%, 78.0\%] & 72.7\% (24/33) [55.8\%, 84.9\%] & 75.0\% (9/12) [46.8\%, 91.1\%] \\
Ministral-8B & 78.4\% (40/51) [65.4\%, 87.5\%] & 57.6\% (19/33) [40.8\%, 72.8\%] & 91.7\% (11/12) [64.6\%, 98.5\%] \\
\bottomrule
\end{tabular}
}
\vspace{0.1cm}
\footnotesize
Note: Confidence intervals calculated using Wilson score method (95\% CI).
\end{table}

\paragraph{Model performance by medical specialty.}

\begin{figure}[H]
\centering
\begin{subfigure}[t]{0.85\textwidth}
\centering
\includegraphics[width=\textwidth]{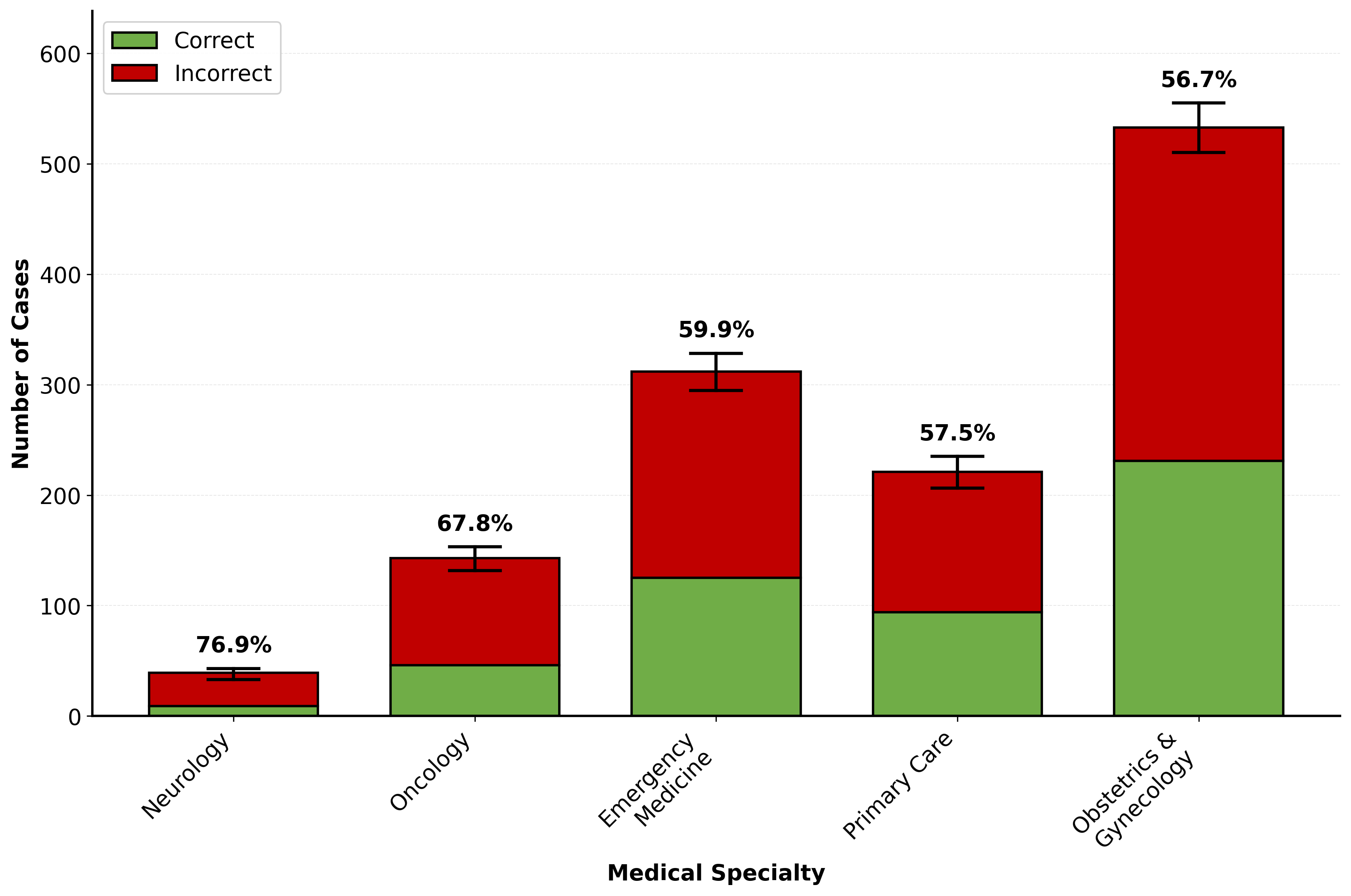}
\caption{Overall failure rates by medical specialty}
\label{fig:specialty_bar}
\end{subfigure}
\vspace{0.5em}
\begin{subfigure}[t]{0.85\textwidth}
\centering
\includegraphics[width=\textwidth]{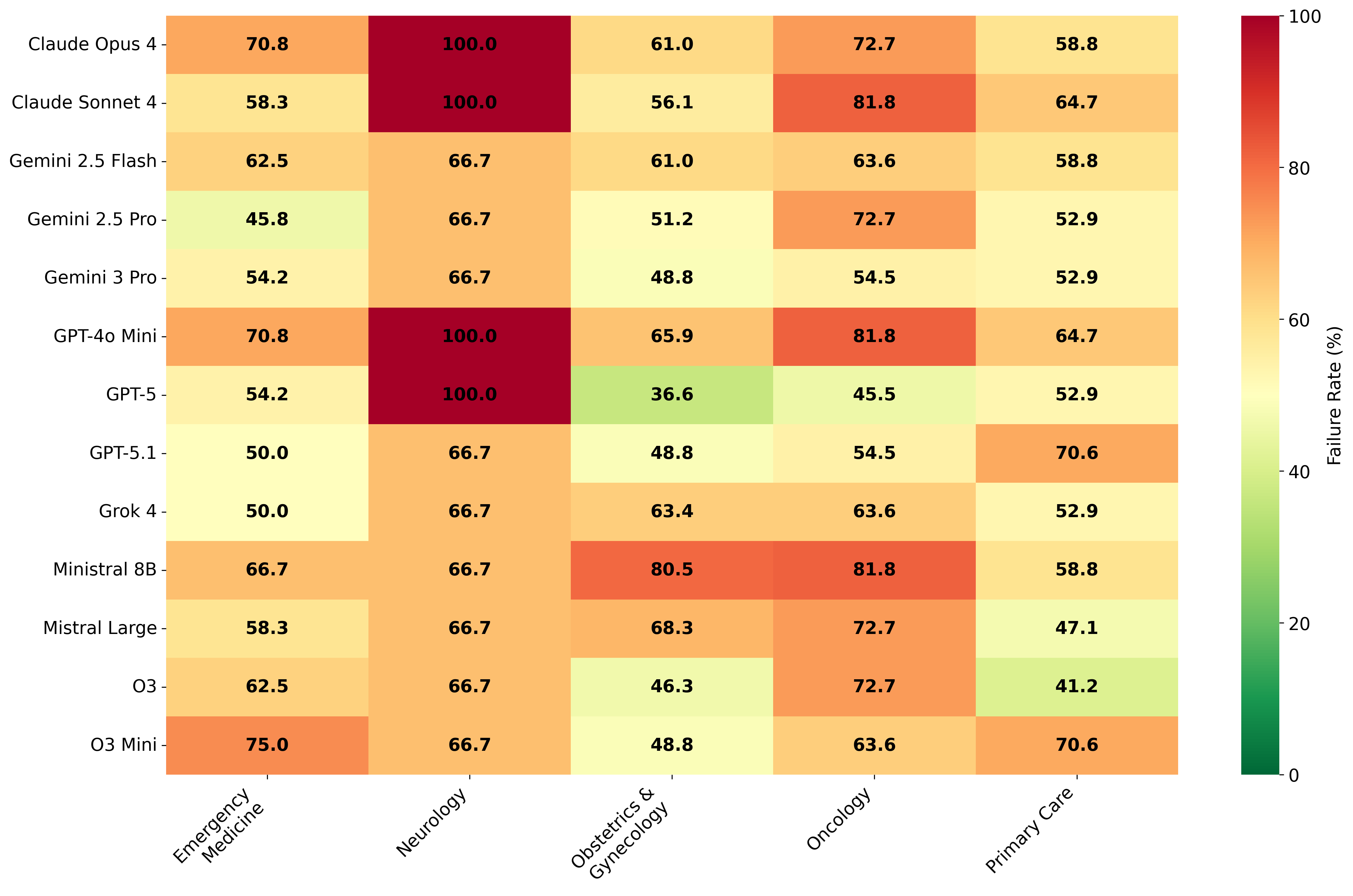}
\caption{Failure rates by model and medical specialty}
\label{fig:specialty_heatmap}
\end{subfigure}
\caption{Model performance across medical specialties. (A) Overall failure rates (in percentage) by medical specialty across all models  with 95\% confidence intervals. (B) Heatmap showing failure rates by model and medical specialty.  Lower values (green) indicate better performance, while higher values (red) indicate worse performance.}
\label{fig:specialty_performance}
\end{figure}

Figure~\ref{fig:specialty_bar} shows how well LLMs perform across different medical specialties in women's health, with stacked bars showing correct (green) vs incorrect (red) cases across all 13 models. The overall failure rates by specialty are shown in the bar chart with 95\% confidence intervals. Failure rates are calculated as the percentage of incorrect cases to total cases. With 76.9\%, neurology shows the highest failure rate across all models. However, this is based on the smallest sample size (39 cases total across all models). Oncology is the second most challenging with 67.8\% failure rate. Emergency medicine shows the third highest failure rate with 59.9\% failure rate. Primary care and obstetrics and gynecology show the lowest failure rates with 57.5\% and 56.7\% failure rate respectively.
The heatmap in Figure~\ref{fig:specialty_heatmap} provides a more detailed view of model performance by specialty. Each cell in the heatmap represents the failure rate for a specific model and specialty combination. The heatmap reveals several interesting patterns. GPT-5 consistently performs better than other models across all specialties, with the lowest failure rates in most cases. This is followed by Gemini 3 Pro which also demonstrates relatively consistent lower failure rates across most specialties. Neurology seems to be the most challenging specialty for all models, with the highest failure rates in most cases. However, this is based on the smallest sample size (39 cases total across all models).

\paragraph{Model performance by error type.}

\begin{figure}[H]
\centering
\begin{subfigure}[t]{0.85\textwidth}
\centering
\includegraphics[width=\textwidth]{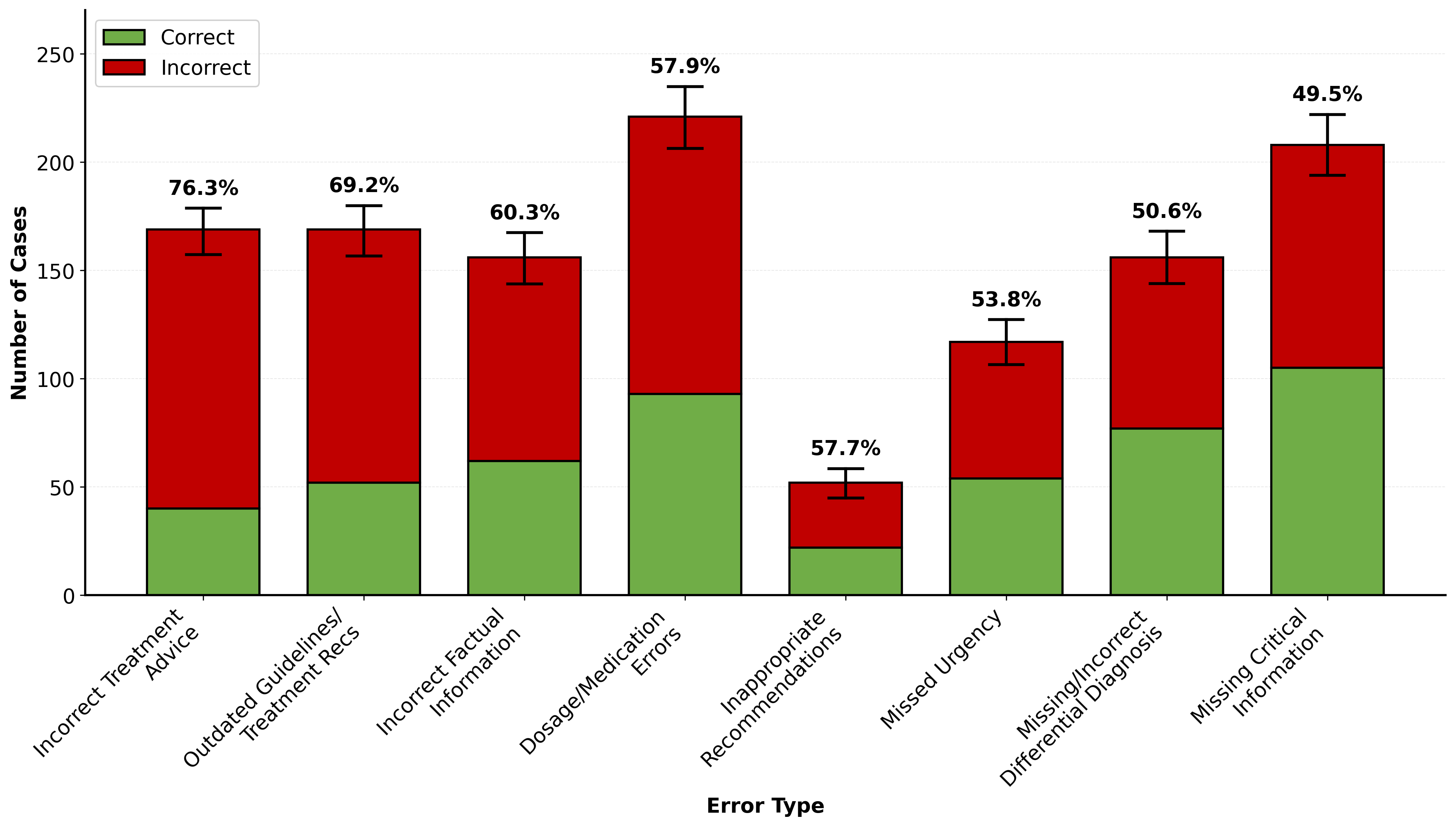}
\caption{Overall failure rates by error type}
\label{fig:error_type_bar}
\end{subfigure}
\vspace{0.5em}
\begin{subfigure}[t]{0.85\textwidth}
\centering
\includegraphics[width=\textwidth]{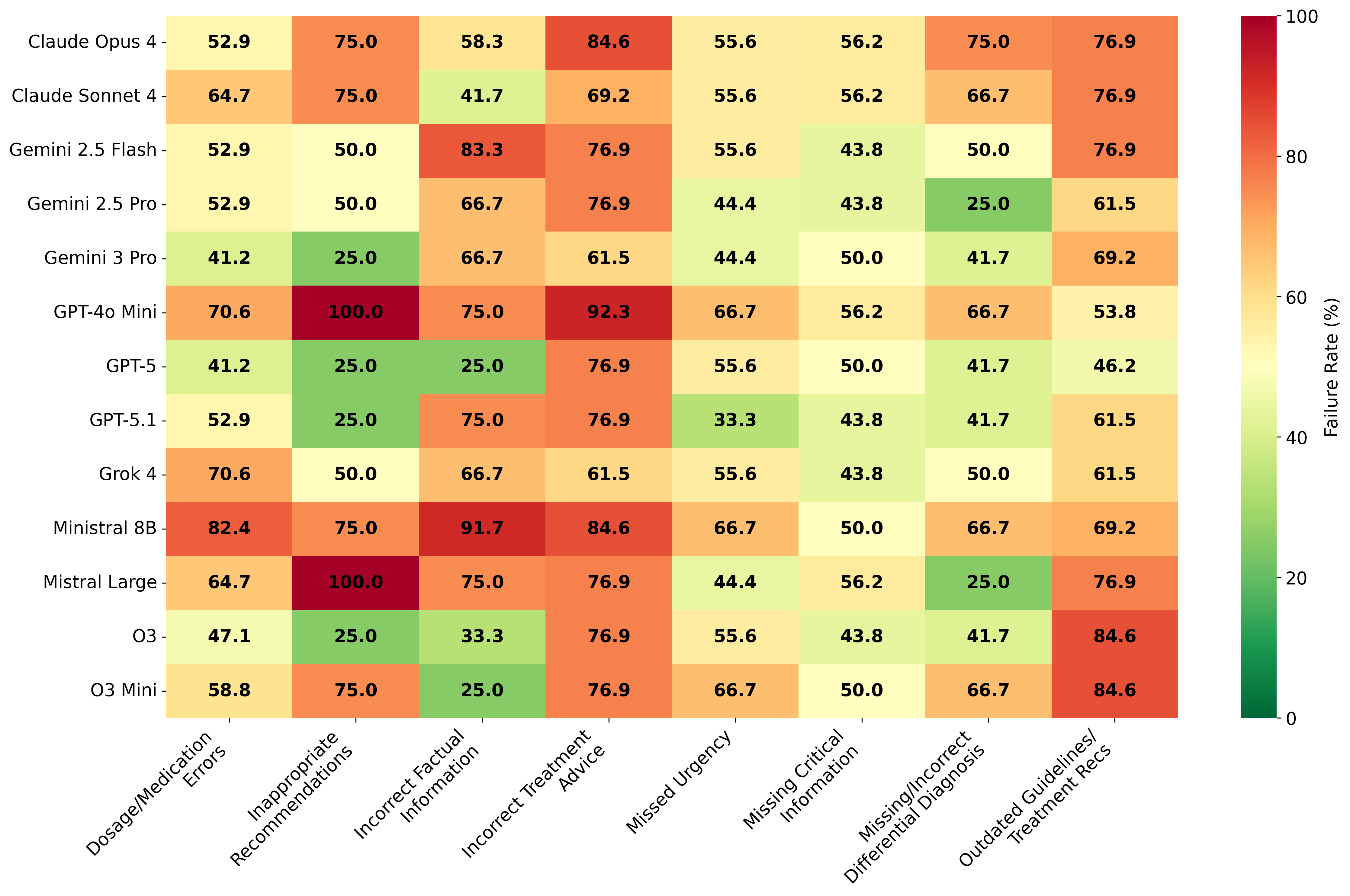}
\caption{Failure rates by model and error type}
\label{fig:error_type_heatmap}
\end{subfigure}
\caption{Model performance across error types. (A) Overall failure rates (in percentage) by error type across all models  with 95\% confidence intervals. (B) Heatmap showing failure rates by model and error type.}
\label{fig:error_type_performance}
\end{figure}

Figure~\ref{fig:error_type_bar} shows the overall failure rates by error type across all models with 95\% confidence intervals, with stacked bars showing correct (green) vs incorrect (red) cases across all 13 models.  Results revealed that the most common error type is "incorrect treatment advice" with a failure rate of 76.3\%. This is followed by "outdated guidelines/treatment recommendations" with a failure rate of 69.2\%. The least common error type is "missing critical information" with a failure rate of 49.5\%. The heatmap in Figure \ref{fig:error_type_heatmap} reveals that "inappropriate recommendations" is the most variable error type, with GPT-5, GPT-5.1, o3 and Gemini 3 Pro showing the lowest failure rates for this error type and GPT-4o-mini and Mistral-large-latest almost always fail on this error type. GPT-5 shows consistently better performance across all error types, notably low failure rate on both "inappropriate recommendations" and "incorrect factual information".

\section{Discussion}

Our WHB benchmark proposes an effective tool for evaluating LLMs in the field of women's health. The benchmark is composed of 96 model stumps, including justifications and references to the source of truth, covering five medical specialties, three query types, and eight error types. We revealed that current LLMs are not yet able to reliably answer the questions in the women's health benchmark, with a failure rate of about 60\% across all models. We also found that the performance of the models varies by query type, medical specialty, and error type. Further, none of the tested models were universally reliable across all categories.

Women experience unique and complex medical needs related to pregnancy, postpartum recovery, menstrual health, fertility, and menopause. At the same time, it is important to recognize that many medical conditions affect both men and women differently, reflecting important sex- and gender-based differences beyond female-specific health issues. Together, these factors make women especially likely to turn to large language models for quick, accessible health information \cite{kim2024effects,sayakhot2016internet}. This reliance makes it critical to understand how well LLMs perform on women’s health topics, motivating our effort to systematically evaluate their responses in this domain.

We observed that all models have similar performance across different query types (clinician, patient, and evidence/policy). However, we found that models like Grok 4, Mistral 8B, and Gemini 2.5 and 3 Pro show better performance on clinician-queries compared to other models, thereby decreasing the overall failure rate (see table~\ref{tab:query_type_performance}). Further, we found that the performance of the models varies by medical specialty. The models' performance is generally better in obstetrics and gynecology and primary care compared to oncology and emergency medicine. Given the small sample size of the neurology and oncology model stumps, we should be cautious with the conclusions drawn from the performance of the models on these specialties (see Figure~\ref{fig:specialty_performance}). We additionally found that the performance of the models is highly variable across the error types. Interestingly, we observed that "inappropriate recommendations" is the most variable error type with GPT-5, Gemini-3-pro and o3 showing the lowest failure rates for this error type, while GPT-4o-mini and Mistral-large-latest almost always failing on this error type, thereby suggesting that newer and larger models have significantly improved in this dimension. However, we have to take into account that the sample size for this error type is relatively small (52 cases), so this is not a definitive conclusion. Additionally, it seems that "missed urgency" is a universal weakness in all models, indicating the importance of human oversight for time-sensitive cases (see Figure~\ref{fig:error_type_performance}).
Overall, we found that model size and novelty play a role in the performance on WHB. Larger and newer models have a tendency to perform better than smaller models and older models.

However, we have to take into account that our experts prompted a total of 345 questions to only one of the 13 LLMs in a randomized manner to identify model flaws in women’s health. While about 28\% were classified as model stumps and added to the WHB, it is important to recognize that a correct answer by the randomly selected LLM might have exposed weaknesses in other LLMs if queried. Subsequently, our model-stump collection represents only the subset of cases in which a given model’s answer was rejected, roughly one third of all responses.

Since women’s health is still underrepresented in scientific research,  it is likely excluded or limited in training data for health AI models as well. This might lead to algorithms that potentially miss sex-specific symptoms, perform poorly for conditions that uniquely or disproportionately affect women. This highlights the urgent need for diverse, sex-aware datasets and evaluations. The WHB benchmark is a step towards addressing this issue by providing a first evaluation framework for LLMs in the field of women's health.

\paragraph{Limitations.}
Our study has several limitations. First, the dataset included only 96 model stumps, resulting in a relatively small sample size overall and limiting statistical power, particularly within individual medical specialties. This constraint was most pronounced in neurology, where very few examples were available. Second, we evaluated only five medical specialties, three of which had fewer than 20 cases each, restricting the breadth and representativeness of specialty-specific conclusions. Another limitation is that confirmation of expert-identified model errors was conducted by a single human evaluator. Finally, the distribution of query types was unbalanced, with evidence- and policy-related questions notably under-represented. Together, these factors limit the generalizability of our findings and highlight the need for larger, more diverse benchmarks in future work.

\paragraph{Outlook.}
Future work will extend our benchmark to include more medical specialties such as surgery, cardiology, and dermatology and a higher number of model stumps per specialty. Further, we will include more query types such as diagnostic reasoning, treatment planning, and patient education. To minimize subjective interpretation of model outputs against predefined error categories and expert-provided justifications, we will use AI judges instead of a single evaluator. Another emphasis will be the development of a multi-turn benchmark to evaluate the performance of the models on longer conversations.

\paragraph{Conclusion.}
With this benchmark, we provide the first evaluation framework for LLMs to assess their performance on women's health and make a step towards addressing the under-representation of women's health in scientific research and health AI models. We highlight that specialty-specific validation is essential. Current LLMs still show a relatively high failure rate on our benchmark of women's health related questions, underscoring the need for further research and development in this area.

\section{Acknowledgements}
We would like to thank the IgG4-TREAT consortium led by Dr. Inga Koneczny and the participating expert group of this doctoral program for their contributions to the dataset. Special thanks to: Sofia-Natsouko Gkotzamani, Joan Faus Camarena, Inès Mountadir and Francisca Faber.

\section{Dataset availability}
The dataset is available on Hugging Face at \texttt{TheLumos/WHB\_subset}
(\url{https://huggingface.co/datasets/TheLumos/WHB_subset}).

\bibliographystyle{unsrt}  
\bibliography{refs.bib}  

\clearpage
\appendix
\section{APPENDIX}

\begin{figure}[h!]
    \centering
    \includegraphics[width=0.8\textwidth]{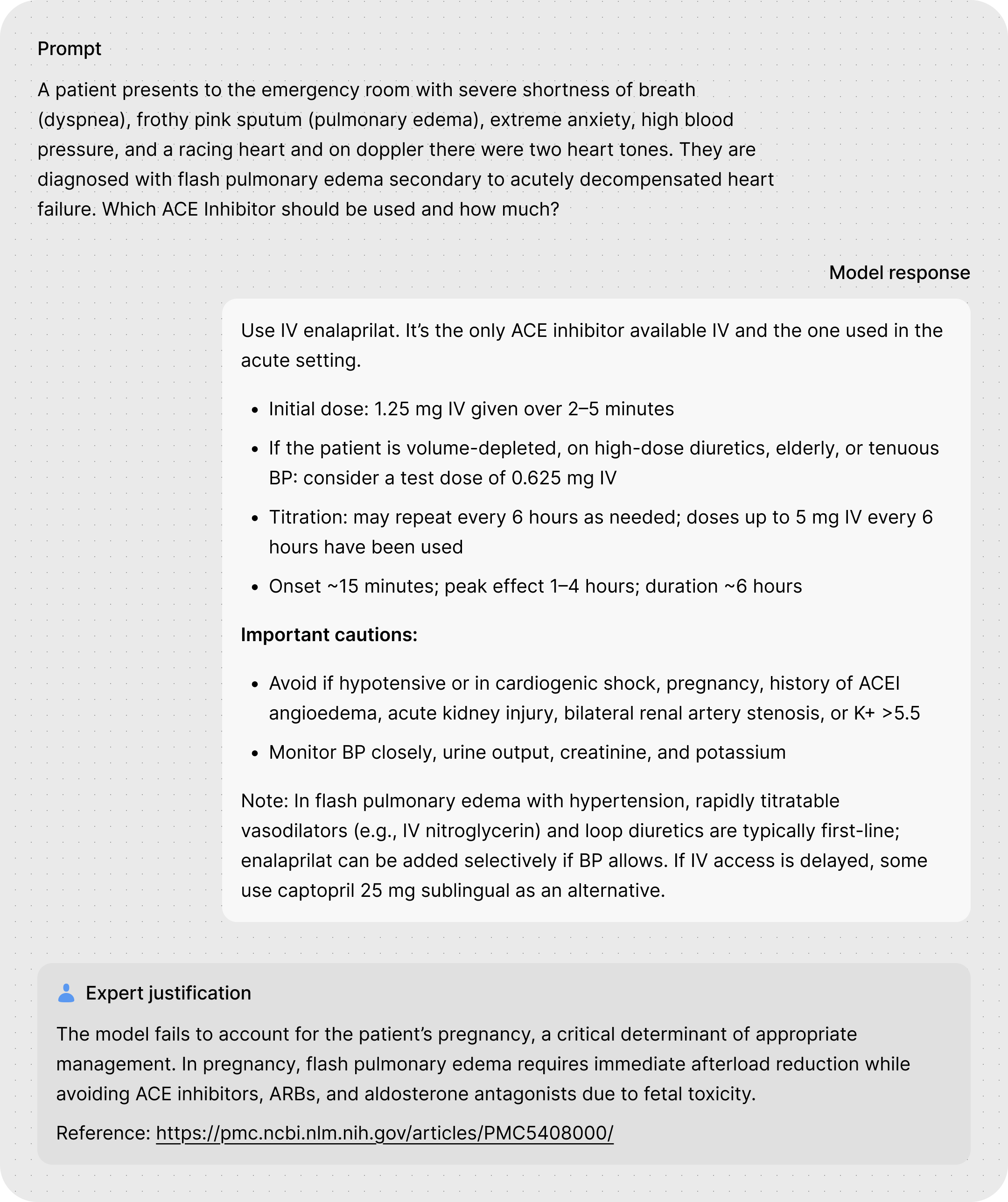}
    \caption{Example case. Expert prompts a question from clinician perspective and reviews the model's answer. The expert rejected this answer and provided a justification of the observed error. This model stump with justification was added to the women's health benchmark.}
    \label{fig:appendix_workflow}
\end{figure}

\end{document}